# BIOS: An Algorithmically Generated Biomedical Knowledge Graph

Sheng Yu[1,2,*], Zheng Yuan[1,2,#], Jun Xia[3,#], Shengxuan Luo[1,2], Huaiyuan Ying[4], Sihang Zeng[5], Jingyi Ren[3], Hongyi Yuan[1,2], Zhengyun Zhao[1,2], Yucong Lin[6], Keming Lu[7], Jing Wang[3], Yutao Xie[3], Heung-Yeung Shum[3]

[1]Center for Statistical Science, Tsinghua University, Beijing, China;

[2]Department of Industrial Engineering, Tsinghua University, Beijing, China;

[3]International Digital Economy Academy, Shenzhen, China;

[4]Department of Mathematical Sciences, Tsinghua University, Beijing, China;

[5]Department of Electronic Engineering, Tsinghua University, Beijing, China;

[6]Institute of Engineering Medicine, Beijing Institute of Technology, Beijing, China;

[7]Department of Industrial & System Engineering, University of Southern California, Los Angeles, CA, USA.

*Correspondence to: syu@tsinghua.edu.cn. #These authors contributed equally.

## ABSTRACT

Biomedical knowledge graphs (BioMedKGs) are essential infrastructures for biomedical and healthcare big data and artificial intelligence (AI), facilitating natural language processing, model development, and data exchange. For decades, these knowledge graphs have been developed via expert curation; however, this method can no longer keep up with today's AI development, and a transition to algorithmically generated BioMedKGs is necessary. In this work, we introduce the Biomedical Informatics Ontology System (BIOS), the first large-scale publicly available BioMedKG generated completely by machine learning algorithms. BIOS currently contains 4.1 million concepts, 7.4 million terms in two languages, and 7.3 million relation triplets. We present the methodology for developing BIOS, including the curation of raw biomedical terms, computational identification of synonymous terms and aggregation of these terms to create concept nodes, semantic type classification of the concepts, relation identification, and biomedical machine translation. We provide statistics on the current BIOS content and perform preliminary assessments of term quality, synonym grouping, and relation extraction. The results suggest that machine learning-based BioMedKG development is a viable alternative to traditional expert curation.

# 1. INTRODUCTION

Biomedical knowledge graphs (BioMedKGs) are specialized databases for the formal representation of biomedical data and knowledge. In BioMedKGs, biomedical concepts are represented as graph nodes, and the relationships between concepts are represented as graph edges. BioMedKGs are the most important informatic infrastructure for biomedical big data and artificial intelligence (AI) because of three essential components. The first component is the terms (names) of each concept, which are synonymous with each other and are considered part of the node properties. The terms can be formal or informal and are intended to include as many variations as possible. For example, type 2 diabetes, type II diabetes, type 2 diabetes mellitus, T2DM, non-insulin-dependent diabetes, NIDDM, and so on, all refer to the same concept and belong to the same node. The terms are critical for natural language processing (NLP), which is used to recognize mentions of biomedical concepts by any of their names in free text, such as electronic health records (EHRs), research papers, and conversations between doctors and patients[1,2]. The second component is the relations (the graph edges). The relations are typed and directed (some can be bidirectional). By connecting two concepts, the relations form a triplet, such as [*Acetaminophen*, *may treat*, *Fever*], where "may treat" is the relation and *Acetaminophen* and *Fever* are the connected concepts and are referred to as the head entity and tail entity, respectively. Relations are critical information for numerous AI tasks, such as automatic diagnosis[3], question-answering[4], and drug discovery[5,6]. The third essential component of BioMedKGs is the ID system, which is used for standardized representation. For example, a concept ID represents the same concept regardless of which term or language is used, facilitating interoperability and data exchange between systems, institutions, and countries.

The development of BioMedKGs requires a tremendous amount of expert input and is extremely expensive[7]. As a result, BioMedKGs developed from scratch are usually limited in size and constructed over decades. Some BioMedKGs have been built on top of existing ones to unify them for focused domains. For example, the Human Phenotype Ontology is built on top of OMIM, Orphanet, and DECIPHER and focuses on phenotype-driven differential diagnostics, genomic diagnostics, and translational research[8]. Some projects have used aggregation to create massive BioMedKGs. The largest BioMedKG is the Unified Medical Language System (UMLS)[9,10], a long-term project that was started in 1986. More than 200 vocabularies and BioMedKGs were aggregated to develop the UMLS, which contains 4.5 million concepts and 14 million terms in 25 languages as of Release 2021AB. The UMLS also contains 21 million relation triplets labeled with 974 different types of relations. However, despite the large size, users often find that the UMLS does not include enough concepts, terms, or relation triplets to meet the growing demand for NLP and AI development. Because of the decades-long development that integrated vocabularies with even longer histories, it is hard to imagine that any expert-curated BioMedKG could be larger than the UMLS, which might have reached the limit for human expert curation. Recently, deep learning-based NLP models have become increasingly accurate to the point that they can be deployed for real-world services. As NLP technology evolves and approaches human parity for many tasks, we believe that the transition from expert-curated BioMedKGs to algorithmically generated BioMedKGs has become feasible.

In this work, we introduce the development of the Biomedical Informatics Ontology System (BIOS), which, to our knowledge, is the first large-scale publicly available BioMedKG fully generated by machine learning algorithms. The content of the current BIOS release was learned from PubMed abstracts and PubMed Central articles. It contains 4.1 million concepts, 7.4 million terms, and 7.3 million relation triplets. Unlike previous work that has used machine learning to address a particular part of BioMedKG construction, BIOS uses machine learning throughout the BioMedKG development process, including (1) the curation of biomedical terms, (2) the aggregation of synonyms to create concept nodes, (3) semantic type classification of the concepts, and (4) relation identification, as illustrated in Figure 1: Development of BIOS involves the full BioMedKG construction process. The steps are introduced in the Methods section. Specifically, a-c: Section 3.1, d-g: Section 3.2, h: Section 3.3, i: Section 3.4, j: Section 3.5, and k-n: Section 3.6.
. Furthermore, the goal is for BIOS to be a multilanguage BioMedKG. Thus, at this initial stage, we translate the English terms to Chinese, a low-resource language in medical informatics, using biomedical machine translation. BIOS can be accessed and downloaded at https://bios.idea.edu.cn. The design and development of BIOS is focused on healthcare (e.g., diagnosis and treatment) instead of drug development or biochemistry. We hope that the release of BIOS and its future updates can serve as a useful AI and big data infrastructure in the healthcare industry and the research community.

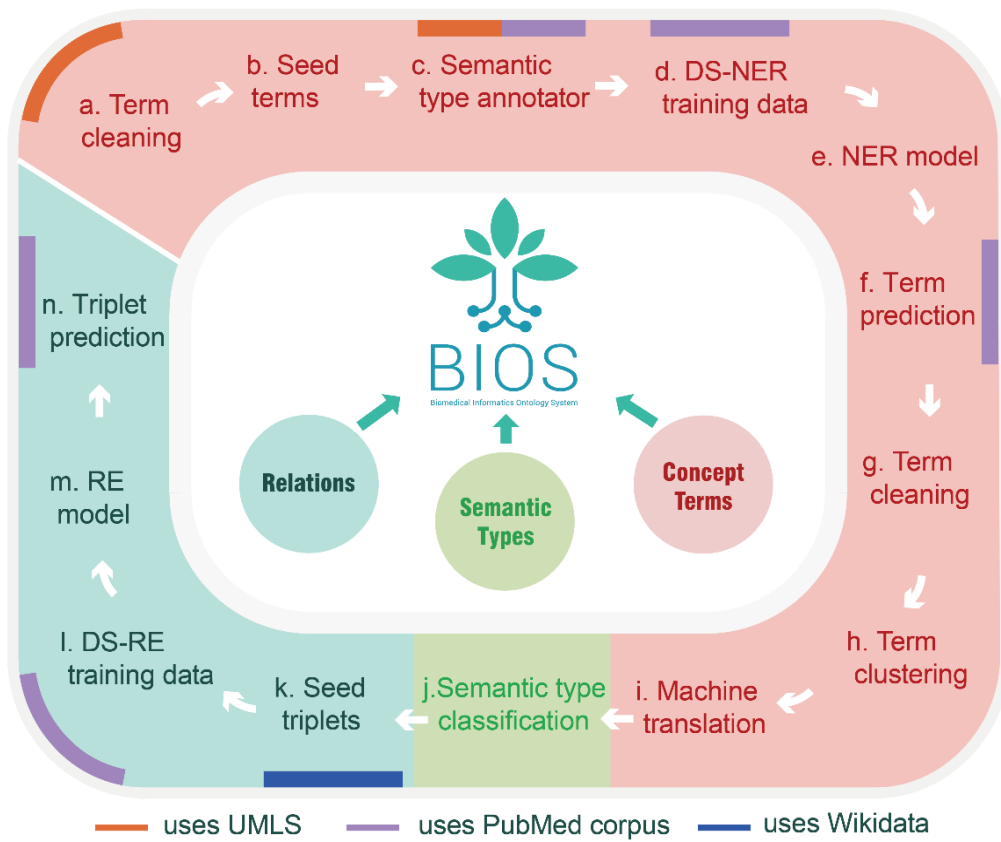

Figure 1: Development of BIOS involves the full BioMedKG construction process. The steps are introduced in the Methods section. Specifically, a-c: Section 3.1, d-g: Section 3.2, h: Section 3.3, i: Section 3.4, j: Section 3.5, and k-n: Section 3.6.

## 2. RELATED WORK

### 2.1 Related work on building BioMedKGs

We review general BioMedKGs (instead of those for specific domains, such as cancer) that were developed using machine learning and NLP. Previous work on machine learning-based integration, inference, or refinement of existing BioMedKGs is not reviewed, as these works did not generate new concepts or terms and were less effective for inferring new relations.

Several existing approaches are based on EHR data. An early work from Finlayson et al. identified mentions of biomedical concepts based on existing ontologies and applied a series of statistical NLP techniques for data cleaning; they reported the cooccurrence data of the identified concepts, which formed a graph, but not in the sense of a BioMedKG, which requires directed and typed relations[11]. Rotmensch et al. constructed a knowledge graph according to EHR data by using the manually curated Google Health Knowledge Graph and the UMLS as dictionaries to identify disease and symptom concepts from EHR free text and learned their associations using the logistic regression, naïve Bayes, and Bayesian network classification models[12]. Li et al. applied the BiLSTM-CRF model[13] to identify symptoms in clinical notes (other entities were extracted from structured data). However, they reported that the model had generalizability issues, which is a common phenomenon that we address in this paper. Additionally, synonym grouping was achieved by using a mapping dictionary, i.e., the concepts were predefined instead of learned. Their relation extraction method was also based on cooccurrence data rather than sentence meaning[14].

Similar to our approach, SemMedDB[15,16] and KGen[17] used PubMed as the corpus to construct knowledge graphs. Because SemMedDB identified UMLS concepts and relations with statistical NLP, it could not generate concepts and terms that were not in the UMLS. KGen applied a different set of statistical NLP methods. Similar to SemMedDB, the identified terms were linked to UMLS concepts, i.e., no new concepts were generated. However, unlike SemMedDB, the verbs of the sentences were used as the relations in KGen, i.e., the relations were in the free text form instead of being controlled. In comparison, BIOS does not rely on existing ontologies and instead learns its own terms and concepts, similar to the work of the Never-Ending Language Learner (NELL)[18,19]. However, BIOS does not reuse predictions as new training samples, which is a hallmark of NELL. BIOS emphasizes techniques of grouping synonyms to form concepts, which is both novel and important for NLP in biomedicine.

**2.2 Related NLP technologies**
We also briefly review recent NLP technologies that are relevant to the construction of BIOS.

Named entity recognition (NER): NER is a sequential tagging task that is used for identifying biomedical terms. Tokens in an input sentence are classified as the beginning token of a term (tagged as B), a non-beginning token of a term (tagged as I), or a token not in the term (tagged as O). By extending the tags with semantic types (a coarse classification of the concept), such as B-diseases, we can achieve NER and semantic type classification simultaneously. The NER model employed in BIOS is the standard BERT sequential classifier[20], which has replaced BiLSTM-CRF as the standard deep learning NER model. Our approach of using automated annotations to generate training data is a type of distant supervision NER (DS-NER), and recent studies have mainly focused on sample cleaning techniques[21–23]. Recent novel developments in discontinuous NER[24,25], nested NER[26–28], and Seq2Seq NER[29] have not yet been employed in the BIOS project due to concerns about engineering maturity.

Biomedical term embedding (BTE): Similar to word embedding[30,31], BTE aims to embed terms with real-valued vectors that should be similar (e.g., by the cosine similarity) when the terms are related or have similar meanings. However, unlike word embedding, BTE should be able to embed any given term, including incorrectly spelled terms, and should not be impacted by the out-of-vocabulary problem. BTE is a critical tool in BIOS for grouping synonyms to form concepts. BTE has also been used in relation extraction and in assessing machine translation quality. SapBERT[32] and CODER[33] are two recent state-of-the-art BTE models based on contrastive learning. The former is trained according to the synonym structure of the UMLS, and the latter is trained according to the UMLS synonyms and relations. However, neither model can embed terms meaningfully to the point where synonymous terms can be grouped as concepts by clustering techniques. Therefore, we developed CODER++, which will be introduced in the Methods section. Classification is another potential approach for synonym clustering[34], but BTE was adopted due to the flexibility of the embedding vectors.

Relation extraction (RE): RE is an NLP classification task that predicts whether a sentence expresses a particular relationship between two identified named entities. Many RE models have been proposed for the few available expert-annotated datasets[35–37]. However, the quantity of annotated samples, the number of annotated relation types, and the format of the annotated document are insufficient for training contemporary deep learning models for large-scale RE to construct BioMedKGs. Similar to DS-NER, distantly supervised RE (DS-RE)[38] has become popular for addressing the training sample problem to achieve high-throughput relation extraction. Recent research on DS-RE has focused on improving the quality of the labels by filtering noisy samples, adjusting sample weights, and generating realistic samples[39–45]. For engineering robustness and modularization, models that jointly detect named entities and relations[46–48] are not considered currently for constructing BIOS.

Biomedical machine translation (BMT): Contemporary deep learning-based machine translation (MT), especially the Transformer architecture[49], can offer satisfactory translation quality. However, the required training set of tens of millions of parallel sentences is prohibitive for the biomedical domain and low-resource languages. The BMT model adopted in BIOS was trained with samples acquired by a novel sentence alignment model using parallel documents[50]. While research on unsupervised MT is promising [51–53], thus far, it is still experimental and lacking in engineering maturity.

## 3. METHODS

### 3.1 Annotation for NER

To identify terms discovery with DS-NER, we need to use automatic methods to annotate a corpus with existing ontologies, i.e., tagging each token with B, I, or O along with the term's semantic type classification, to provide a training sample. We manually selected 8 vocabulary sources from the UMLS (Release 2020AB) that we found provided adequate coverage of biomedical terms. At this initial stage, we included only 64 semantic types from the UMLS, and additional important semantic types, such as genes, will be added in the next release.

The UMLS contains a large number of terms that are not biomedical terms according to common standards, as illustrated by the example in the Results section. Since the main annotation method is string matching using the forward maximum matching algorithm (FMM), to reduce the possibility of matching strings that are spelled the same as recorded terms but have different meanings, for each vocabulary source, we filtered its terms with manually selected term types specific to that source. For instance, permutations and acronyms were not included. Finally, extensive filtering rules based on dictionaries, regular expressions, and sentence parsing were applied to remove common words and problematic terms that may cause errors during FMM.

The remaining terms, which are referred to as "seed terms", can still be ambiguous. In particular, a term can have multiple biomedical meanings with different semantic types. Therefore, we trained a dedicated semantic type annotator (STA) to predict the semantic type of a term using both the term itself and its surrounding text as input. The training samples were multiword terms from the UMLS that in general, do not have ambiguous daily meanings. The semantic types of these terms were used as the label, and if a term had multiple semantic types in the UMLS, a random term was used based on the fact that a large sample size can overcome moderate noise in the data. The classification model was trained on PubMedBERT[54], a BERT model that was pretrained on PubMed abstracts.

With the STA, we performed automatic annotation as follows: We used the seed terms as a dictionary to perform FMM on the PubMed abstracts and half of the PubMed Central full texts (referred to as the PubMed corpus hereinafter). For each matched term, we applied the STA to predict its semantic type. If the predicted type belonged to the same semantic group (a UMLS hierarchy for semantic types) as any of the recorded semantic types of that term in the UMLS, the term was annotated with that recorded semantic type with a probability of $1 - DF$, where $DF$ is the fraction of documents that contain the term; otherwise, the matched term was considered as an incorrect match and was not annotated.

**3.2 Term discovery and cleaning**
We used PubMedBERT to train a B-I-O sequential tagging model for NER. As reported in Lin et al.[14], machine learning-based NER models tend to memorize annotated terms in the training data and have limited generalizability for identifying terms that they have not encountered. This phenomenon is even more prominent with deep learning models because they have a much larger parameter space for memorizing than conventional models. With the ability to automatically annotate very large datasets, we addressed this issue by using a different sampling strategy. Specifically, instead of annotating a corpus that may contain repeated mentions of only some of the seed terms, we sampled only one sentence per term in the PubMed corpus. This sampling strategy covered more terms than the traditional method. More importantly, the model was presented with only one occurrence per term. In other words, the model was repeatedly presented with the patterns around the terms, and it attempted to memorize those surrounding patterns instead of the terms. Thus, the model was generalizable and could discover new terms.

The rules for cleaning the seed terms were also applied to clean the discovered terms. In addition, we further cleaned the data by performing FMM using the newly discovered terms as the dictionary on the PubMed corpus and filtered a term if the ratio between its FMM count in the corpus and the number of times it was predicted by the NER model was greater than a certain threshold, which indicated that the term was only predicted occasionally and was most likely an error. Finally, we filtered terms that were not noun phrases or did not end with nouns in the sentence where they were predicted, which was based on the Stanford Parser[55].

### 3.3 Synonym grouping to form concepts

Thus far, we have identified only individual terms that still need to be linked to corresponding concepts. Intuitively, one might think that this could be achieved by mapping terms to concepts as a classification task. However, as BIOS is built entirely from scratch, there is no predefined concept set. In other words, the concepts need to be computationally defined based on the discovered terms. Our strategy relied on the similarity of the BTE vectors to identify synonyms, and each synonym group was defined as a concept. This technique required that the BTE model meaningfully understood any given biomedical term and embedded synonymous terms much closer than nonsynonymous terms, and no existing BTE model satisfied these requirements. In our experiments, we observed that CODER and SapBERT performed poorly in terms of understanding small differences, such as the differences between 1 and 2 or between α and β. To improve the model's ability to distinguish close terms, we continued training CODER with difficult samples: each sample was a UMLS term accompanied by the 30 terms closest in the embedding, which were retrieved by Faiss[56], and the labels were whether the terms were synonyms in the UMLS. Faiss's indexing was periodically updated to reflect the most recent state of the BTE model. We referred to the new model as CODER++[57]. The output of CODER++ was 768-dimensional vectors. We manually selected a conservative similarity threshold of 0.8 for identifying synonyms, which yielded a very high precision and moderate recall.

We then clustered the identified terms as follows. First, we considered all of the terms as nodes of an undirected graph, with the edge weights determined by the BTE cosine similarity of the connected terms. Next, edges with weights less than 0.8 were removed, and the graph became a collection of connected subgraphs. Intuitively, terms in the same subgraph were synonymous; however, some subgraphs contained hundreds of terms. Therefore, for subgraphs with more than 50 terms, we performed recursive graph bipartition with Ratio Cut[58] until the subgraph had fewer than 50 nodes or the cosine similarity of the mean BTEs of the two subgraphs was greater than 0.6. This concluded the term clustering task, and each subgraph was treated as a biomedical concept.

**3.4 Biomedical machine translation**
BIOS adopted Luo et al.'s BMT model, which reported a remarkable BLEU score of 35.04 for English-Chinese translation and 40.13 for Chinese-English translation[50]. According to our experiments, the model could translate common terms accurately because it had memorized these terms based on the training data. However, the model could translate uncommon terms, such as complicated chemical names, very arbitrarily. We used a back-translation method to automatically determine which translations were unreliable. We used BMT to translate each term from English to Chinese, then translated the term back to English. The Chinese translation was considered unreliable and would be deleted if the CODER++ similarity of the original and back-translated English terms was less than 0.55; chemicals required a higher threshold of 0.8. This process was based on the intuition that if the Chinese text was generated arbitrarily, the back-translated English text would not be similar to the original term. The thresholds were selected empirically.

**3.5 Semantic type classification**
The semantic type schema of BIOS was modified from that of the UMLS according to our understanding of relevance and ease of use for healthcare big data and AI. For example, we merged all 26 subclasses of chemicals from the UMLS into a single semantic type, *Chemical or Drug*.

To determine the semantic types of the concepts, we aggregated the NER model's predictions. The original predictions included 64 UMLS types, which were mapped to 18 BIOS types. Each concept could have multiple terms; each term could be predicted multiple times in the PubMed corpus by the NER model, and each time, an individual prediction was made. Therefore, we counted the type distribution by aggregating the type predictions of all the terms belonging to the same concept, and we kept the semantic types that reached 1/3 of the total counts. Therefore, theoretically, a concept could have up to 3 semantic types, although most concepts had a single type.

**3.6 Relation extraction**
Similar to NER, we used distant supervision to train the RE model, and we used Wikidata as the label source. Wikidata is a general domain knowledge graph. We performed exact matching on Wikidata page titles (not aliases) using the BIOS terms as a dictionary to identify biomedical concepts, and we required that the page had at least one manually selected biomedical relation for disambiguation. We mapped the selected relations from Wikidata to $K = 19$ BIOS relations to construct the model, which included 9 pairs of unidirectional relations (e.g., "is a" and "reverse is a") and 1 bidirectional relation ("significant drug interaction"). We then retrieved sentences from the PubMed corpus that contained both head and tail entities of Wikidata triplets to form a DS-RE dataset. The entity matching in the sentences was based on FMM using all the BIOS terms, and the two terms were required to be no more than 10 tokens away from each other. Each sample was a bag of sentences with the same pair of head and tail entities, and the label was a $K$-dimensional binary vector $(y_1, \ldots, y_K)$ that indicated which relations were true (multilabel classification). Artificial negative samples ($y_k = 0$, for $k = 1, \ldots, K$) were generated by identifying sentences that contained entities whose semantic types were compatible with any of the $K$ relations but with triplets that were not recorded in Wikidata.

The RE model was trained by the bags of sentences. Each sentence input $x = (x_0, \ldots, x_l)$ was a sequence of tokens. Depending on the relative positions of the head and tail entities, the input could have one of the following forms:

$$(x_0, \ldots, [\text{H\_ST}], x_{hs}, \ldots, x_{he}, [\text{H\_ED}], \ldots, [\text{T\_ST}], x_{ts}, \ldots, x_{te}, [\text{T\_ED}], \ldots, x_l),$$
$$(x_0, \ldots, [\text{T\_ST}], x_{ts}, \ldots, x_{te}, [\text{T\_ED}], \ldots, [\text{H\_ST}], x_{hs}, \ldots, x_{he}, [\text{H\_ED}], \ldots, x_l),$$

where $x_{hs}, \ldots, x_{he}$ and $x_{ts}, \ldots, x_{te}$ are the tokens of the head and tail entity terms, which are surrounded by special tokens that indicate their starts and ends. The sentence was encoded by the PubMedBERT PLM:

$$\mathbf{h}_0, \ldots, \mathbf{h}_l = PLM(x),$$
$$\mathbf{h}^x_{PLM} = concat(\mathbf{h}_{[\text{H\_ST}]}, \mathbf{h}_{[\text{T\_ST}]}).$$

Additionally, we used CODER to provide context independent encoding of the head and tail entities:

$$\mathbf{h}_{CODER} = concat\big(CODER(x_{hs}, \ldots, x_{he}), CODER(x_{ts}, \ldots, x_{te})\big).$$

CODER was trained on the UMLS relations, which had only a minor overlap with the Wikidata relations. Therefore, by using CODER, we could transfer basic knowledge from the UMLS to improve the accuracy of the model, with minimal risk of information leakage. The final feature for a sentence was represented as:

$$\mathbf{h}_x = concat(\mathbf{h}^x_{PLM}, \mathbf{h}_{CODER}).$$

To identify which relation was predicted by which sentence, we used the max operation instead of attention[42] to aggregate the sentences in a bag. For sentences $\{x^1, \ldots, x^q\}$, we obtained their representations $\mathbf{h}_{x^1}, \ldots, \mathbf{h}_{x^q}$. The probability that sentence $x^j$ expressed relation $r_k$ was predicted as

$$p_{x^j, r_i} = \sigma\big(W_{r_k} \mathbf{h}_{x^j} + b_{r_k}\big), j = 1, \ldots, q, \text{ and } k = 1, \ldots, K,$$

where $\sigma(x)$ is the logistic function. For each relation, the sentence with the highest probability was used to calculate the cross-entropy loss for backpropagation:

$$m_k = \arg\max_{1 \le j \le q} p_{x^j, r_k},$$

$$Loss = \sum_{k=1,\ldots,K} \left(-y_k \log p_{x^{m_k}, r_k} - (1-y_k) \log(1 - p_{x^{m_k}, r_k})\right).$$

For the prediction, sentences were input into the RE model individually rather than in bags. A relation $r_k$ was predicted for a pair of head and tail entities if the corresponding probability exceeded 0.5.

### 3.7 Data and version management mechanism

The procedures and models introduced in this section were what generated the initial release version of BIOS. However, algorithmically generated BioMedKGs are a powerful concept not because of a specific algorithm but because of their rapid iteration. At this beginning step, our collection of data sources, data processing procedures, and NLP models are all primitive; they have been iterated and improved by many versions and will continue to improve at a fast rate. Thus, for this to work, we have developed a mechanism for managing data production and version control.

As illustrated in Figure 2, the mechanism has three main purposes: indexing, source tracking, and version control. The primary goal of indexing is to facilitate sample generation. For example, an NER training set can be generated by sampling two sentences per indexed term. Currently, the indices include raw sentence information, such as word and noun phrase indices, and distant supervision information, including named entities and relation triplets. Source tracking applies to the prediction results to ensure that for each predicted term or triplet, we know the sentences and documents from which they were identified and the model by which they were predicted, which is important for both referencing and internal reviews of model qualities. Finally, version control is used to maintain the algorithms, including the trained models, their associated training data and source codes, the dependencies, the sampling strategy used to prepare the training data, and the algorithms used to create the distantly supervised annotations. The development of BIOS has been a trial-and-error, rapid iterative process, with some algorithms updated every few days; thus, the data and version management mechanism has been critical.

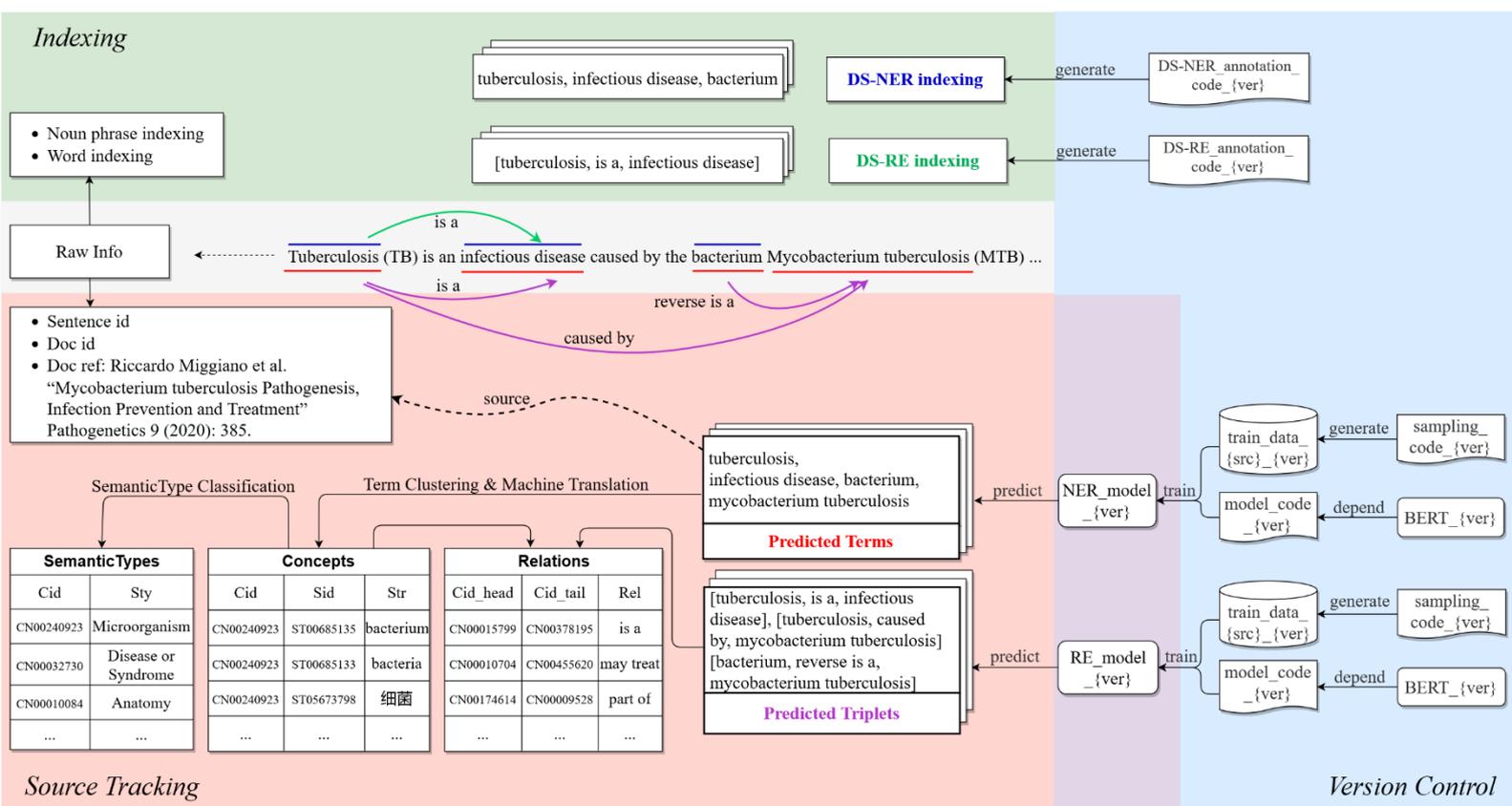

Figure 2: Data production and version control mechanism for the BIOS system. The sentences are indexed to generate training data. The prediction results can track the source text and models. The version control maintains the code and dependencies of each trained model.

## 4. RESULTS

The dictionary of seed terms obtained from cleaning the 8 selected source vocabularies included 1.10 million terms. According to the automatic annotation rule, only 450 thousand seed terms appeared in the PubMed corpus, which could be covered by 440 thousand sentences, following the strategy of sampling 1 sentence per term. The NER model trained with these sentences was applied to predict new terms in the PubMed corpus, yielding a total of 6.10 million unique terms. Among these terms, only 635 thousand (10.4%) were included in the UMLS, reflecting the coverage limitations of the largest expert-curated BioMedKG. After additional term cleaning was applied, we kept 5.20 million English terms, which were then clustered into 4.14 million concepts using CODER++ embedding. Table 1 shows the composition of concepts according to semantic type. Table 2 shows the distribution of the number of terms per concept. FMM was used to determine that the discovered term appeared 612 million times in the PubMed corpus. The term that appeared the most was "cells", which appeared 10.2 million times. However, 35.5% of the terms appeared only once. Table 3 shows the cumulative coverage of the most frequent terms.

Table 1: Composition of concepts in BIOS according to semantic type.

| Semantic type | Count | Proportion |
|---|---:|---:|
| Chemical or Drug | 2,193,599 | 0.507 |
| Disease or Syndrome | 434,196 | 0.100 |
| Therapeutic or Preventive Procedure | 308,836 | 0.071 |
| Anatomy | 198,322 | 0.046 |
| Medical Device | 154,911 | 0.036 |
| Sign, Symptom or Finding | 123,477 | 0.029 |
| Microorganism | 119,015 | 0.028 |
| Neoplastic Process | 120,562 | 0.028 |
| Diagnostic Procedure | 113,385 | 0.026 |
| Laboratory Procedure | 88,709 | 0.021 |
| Physiology | 88,977 | 0.021 |
| Eukaryote | 86,232 | 0.020 |
| Pathology | 86,619 | 0.020 |
| Anatomical Abnormality | 76,633 | 0.018 |
| Mental or Behavioral Dysfunction | 68,191 | 0.016 |
| Injury or Poisoning | 51,226 | 0.012 |
| Research Activity or Technique | 9,556 | 0.002 |
| Research Device | 1,531 | 0.000 |

Table 2: Distribution of the number of terms per concept.

| Terms per concept | Proportion |
|---:|---:|
| 1 | 0.8874 |
| 2-5 | 0.1028 |
| 6-10 | 0.0063 |
| 11-20 | 0.0023 |
| 21-30 | 0.0007 |
| >30 | 0.0005 |

Table 3: Term occurrence coverage of the most frequent terms.

| Number of most frequent terms | Coverage of total occurrences |
|---:|---:|
| 1,000 | 0.460 |
| 10,000 | 0.711 |
| 100,000 | 0.884 |
| 1,000,000 | 0.978 |

Although BIOS (the current release) has 5.2 million English terms and the UMLS (Release 2020AB) has 9.5 million English terms, these two BioMedKGs cannot be compared only on the basis of numbers. The sizes of both databases are "inflated" (BIOS contains many typos that appeared in the PubMed corpus and the UMLS contains many nonnatural language terms such as permutations), and the two vocabularies have drastically different term quality. Figure 3 shows a case study that compares the FMM string matching results obtained using BIOS and the UMLS on the same PubMed abstract[59]. The figure shows that the strings matched using UMLS covered more than half of the text and included many words and numbers that are not biomedical terms. Some of these "terms" were spelling collisions, such as "but" being a term of *butting*, a mental/behavioral dysfunction, and "be" being a term of *bacterial endocarditis*; however, most terms were simply the result of over curation. For inexperienced UMLS users, using all the terms can lead to uninformative results. However, the recall can be reduced by attempting to control the term quality using term types and sources. For example, our NER identified 635 thousand UMLS terms in the PubMed corpus, but after filtering by term types and sources, our seed terms covered only 450 thousand of these terms. In contrast, the strings matched by the BIOS dictionary were all nontrivial biomedical terms, including terms that were not included in the UMLS. There were also missed terms in the BIOS results. For example, the BIOS dictionary only contained "ige-mediated allergic disease" but not its plural form.

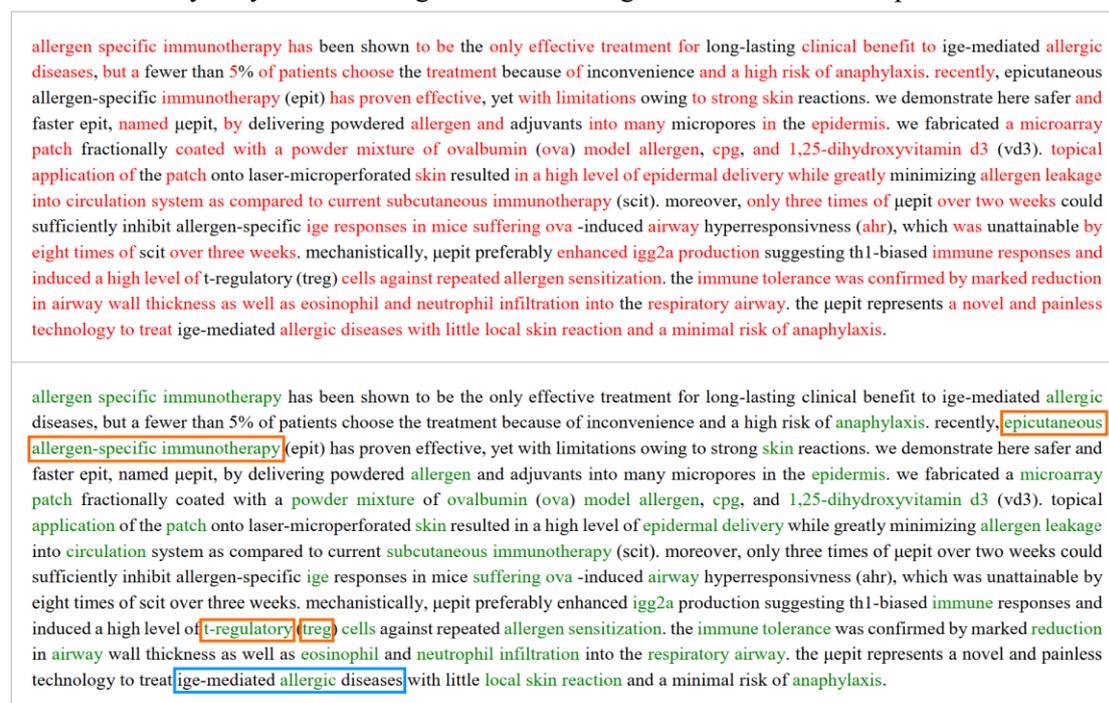

Figure 3: Term matching results using the full UMLS (top) and BIOS (bottom) as dictionaries. A highlighted area may be composed of consecutive matched terms. The orange boxes highlight biomedical terms in BIOS but not in the UMLS. The blue box shows an example term that is missing from the BIOS dictionary.

We evaluated the accuracy of using CODER++ embedding to identify synonymous terms, with the UMLS used as a silver standard. Terms from the intersection of BIOS and the UMLS vocabularies were tested in pairs, and labels were assigned based on whether their concept codes were identical. The classification boundary was whether the embeddings' cosine similarity was greater than 0.8, and the classification achieved a recall of 0.309 and a precision of 0.953. Furthermore, we manually examined the predictions that were ruled as false positives (similarity > 0.8 but with different concept codes in the UMLS) and found that 193 of the 200 randomly sampled "false positives" were synonymous (20 of them are shown in Table 4). Therefore, the corrected precision estimate was 0.998. The use of the UMLS as a silver standard may favor the metrics, as CODER++ was trained using UMLS data. However, the ability to correct the UMLS data suggests that the model is generalizable.

Table 4: Examples of manually reviewed pairs of terms with different concept IDs in the UMLS but with CODER++ embedding similarities greater than 0.8 that were classified as synonyms. Note that many of these terms have multiple UMLS concept IDs, and it is possible that the pair of terms have a common ID. The synonymy evaluation consulted examples from the UMLS. For example, because "cell" and "cell structure" are synonyms in the UMLS, "hepatocyte" and "hepatocyte structure" should also be synonyms.

| Term 1, Concept ID | Term 2, Concept ID | Similarity | Synonym |
|---|---|---|---|
| tumor cell, C0431085 | neoplastic cell, C0597032 | 0.807 | Yes |
| hepatocyte, C0227525 | hepatocyte structure, C0682613 | 0.868 | Yes |
| liver cell, C0227525 | hepatocyte structure, C0682613 | 0.816 | Yes |
| hepatic cells, C0227525 | hepatocyte structure, C0682613 | 0.837 | Yes |
| cerebral hemisphere, C0007783 | brain hemisphere, C0228174 | 0.897 | Yes |
| cerebral hemisphere, C0007783 | hemispherium cerebri, C0228174 | 0.856 | Yes |
| cerebral structure, C0242202 | cerebral, C0228174 | 0.894 | Yes |
| cerebral structure, C0242202 | brain structure, C0006104 | 0.801 | Yes |
| brain tissue, C0440746 | brain tissue structure, C0459385 | 0.956 | Yes |
| synovium, C0039099 | synovial structure, C0584564 | 0.833 | Yes |
| synovial tissue, C0039099 | synovial structure, C0584564 | 0.842 | Yes |
| eyes, C0015392 | set of eyes, C2949813 | 0.902 | Yes |
| ocular structure, C0015392 | set of eyes, C2949813 | 0.815 | Yes |
| eye structure, C0015392 | set of eyes, C2949813 | 0.815 | Yes |
| mucosa, C0026724 | tunica mucosa, C0017136 | 0.821 | Yes |
| clinical decision support systems, C0525070 | clinical decision support system, C4035904 | 0.811 | Yes |
| lhx3, C1416850 | lim homeobox 3, C0300290 | 0.848 | Yes |
| dinoproston, C0012472 | dinoprost, C0012471 | 0.802 | No |
| fmo3, C1414645 | fmo 3, C3888280 | 0.811 | Yes |
| renal disease, C1408247 | renal diseases, C0549526 | 0.911 | Yes |

As an example of synonym grouping in BIOS, Table 5 lists all the terms associated with Concept CN00016530 *Congenital Melanocytic Nevus*. The concept has 14 distinct English terms, with the variations arising mainly from combinations of "melanocytic", "pigmented", or "nevocytic", and "nevi", "nevus", or "naevus". The plural form "congenital melanocytic nevi" was automatically labeled the preferred term because it appeared the most frequently in the

PubMed corpus. Another example is Concept CN00016533 *Systemic Lupus Erythematosus*, which has 60 distinct English terms; however, many of the variations arise from the various ways to spell "erythematosus" incorrectly, such as "erithematosus", "erythmatosus", or "erythemathosus". These typos were not removed because we did not have an automatic method to robustly identify typos and because we thought that keeping the typos could potentially be useful.

Table 5: Terms associated with the concept *Congenital Melanocytic Nevus* in BIOS, as an example of synonym grouping.

| Term | Term type |
|---|---|
| congenital melanocytic nevi | Preferred term |
| congenital melanocytic nevus | Entry term |
| congenital melanocytic naevus | Entry term |
| congenital pigmented nevi | Entry term |
| congenital nevocytic nevi | Entry term |
| congenital nevocellular nevi | Entry term |
| congenital pigmented nevus | Entry term |
| congenital nevocellular nevus | Entry term |
| congenital nevocytic nevus | Entry term |
| congenital pigmented naevus | Entry term |
| congenital dermal melanocytic nevus | Entry term |
| congenital melanotic nevus | Entry term |
| congenital pigmented congenital melanocytic nevus | Entry term |
| congenital pigmentary nevus | Entry term |

A total of 877 thousand relation triplets were identified from Wikidata and directly imported into BIOS. From these triplets, 240 thousand matched sentences from the PubMed corpus, which were used to develop the DS-RE dataset. Table 6 shows the sample size (number of triplets) for each relation type. Each sample was a bag of sentences; the average number of sentences was 42, and we used at most 16 sentences due to GPU memory limitations. The dataset was divided into 100:1:1 partitions for training, validation, and testing, respectively. The precision, recall, and F1-score on the test set are reported in Table 6. The RE model was applied on part of the PubMed corpus to mine the relation triplets, and thus far, 1.10 million triplets have been predicted.

Table 6: Sample size and test set precision, recall, and F1 score of the relation extraction.

|  | Sample size | Precision | Recall | F1-score |
|---|---|---|---|---|
| *All (micro-average)* | 877K | 97.25 | 94.59 | 95.9 |
| *is a* | 170K | 98.82 | 95.57 | 97.17 |
| *part of* | 78K | 97.04 | 88.34 | 92.49 |
| *may treat* | 6K | 90.63 | 93.55 | 92.06 |
| *involved in* | 47K | 91.04 | 98.39 | 94.57 |
| *found in taxon* | 90K | 92.92 | 94.59 | 93.75 |
| *active ingredient in* | 2K | 75 | 60 | 66.67 |
| *expressed in* | 36K | 100 | 95.24 | 97.56 |
| *may cause* | 3K | 78.57 | 91.67 | 84.62 |
| *encoded by* | 5K | 66.67 | 100 | 80 |
| *significant drug interaction* | 2K | 75 | 93.75 | 83.33 |
| *inverse is a* | 170K | 98.06 | 95.88 | 96.96 |
| *reverse part of* | 78K | 96.1 | 90.24 | 93.82 |
| *reverse may treat* | 6K | 97.06 | 98.51 | 97.78 |
| *reverse involved in* | 47K | 96.08 | 87.5 | 91.59 |
| *reverse found in taxon* | 90K | 98.34 | 97.8 | 98.07 |
| *reverse active ingredient in* | 2K | 92.31 | 100 | 96 |
| *reverse expressed in* | 36K | 100 | 99.16 | 99.58 |
| *reverse may cause* | 3K | 100 | 70.37 | 82.61 |
| *reverse encoded by* | 5K | 98.21 | 98.21 | 98.21 |

Finally, in addition to the relations predicted by the RE model and those imported from Wikidata, the NER model was another source of "*is a*" relations. The prediction of the NER model included 67 possible UMLS semantic types, which were simplified to 18 BIOS types. For example, the predicted semantic type for the concept *Tumor Cell* was the UMLS type *Cell*, which was simplified to the BIOS type *Anatomy*. However, the original information was not lost and was converted to corresponding relation triplets, e.g., [*Tumor Cell*, *is a*, *Cell*]. As a result, we obtained 2.86 million "*is a*" relations and an equal number of "*reverse is a*" relations.

BMT was used to translate terms from English to Chinese. Overall, 58.5% of the English terms could be reliably translated to Chinese according to the back-translation rule (including when the Chinese terms appeared the same as in English), and 56.3% concepts had Chinese translations. Table 7 shows the translation rate according to the semantic type. Microorganisms and eukaryotes had the lowest translation rates because the BMT model never saw those words in the training data. In total, we obtained 2.31 million distinct Chinese terms.

Table 7: Concept- and term-wise rates of machine translation.

| **Semantic type** | **Concept** | **Term** |
|---|---|---|
| Anatomical Abnormality | 0.675 | 0.698 |
| Anatomy | 0.676 | 0.691 |
| Chemical or Drug | 0.462 | 0.479 |
| Diagnostic Procedure | 0.716 | 0.730 |
| Disease or Syndrome | 0.798 | 0.814 |
| Eukaryote | 0.114 | 0.114 |
| Injury or Poisoning | 0.692 | 0.734 |
| Laboratory Procedure | 0.771 | 0.780 |
| Medical Device | 0.553 | 0.561 |
| Mental or Behavioral Dysfunction | 0.546 | 0.580 |
| Microorganism | 0.383 | 0.396 |
| Neoplastic Process | 0.849 | 0.861 |
| Pathology | 0.805 | 0.820 |
| Physiology | 0.697 | 0.708 |
| Research Activity or Technique | 0.753 | 0.772 |
| Research Device | 0.680 | 0.676 |
| Sign, Symptom, or Finding | 0.676 | 0.695 |
| Therapeutic or Preventive Procedure | 0.686 | 0.695 |

## 5. DISCUSSION

We started the BIOS project in January 2021, and the first version was made available for download in January 2022. In a year, BIOS has grown to a size comparable to the largest expert-curated BioMedKGs in the world, which have been developed over many decades, and at this rate, BIOS will soon surpass all of these BioMedKGs. While existing BioMedKGs have provided key training materials, proper measures were taken during model training to ensure that BIOS did not simply memorize the training data. As a result, only a small fraction of the BIOS data overlaps with the UMLS data. The BIOS and UMLS data also have significantly different qualities. For example, the terms in BIOS are based on terms that people use in writing, including typos, and BIOS performs better than the UMLS when used as dictionaries for matching biomedical terms. During development, when we tried to use existing BioMedKGs as training or test data, we also found that expert-curated data were less accurate than expected. All of these factors strengthen our confidence in the viability of data-driven and algorithmically generated BioMedKGs.

The development of the BIOS project has many unresolved issues. DS-NER and DS-RE are still immature techniques that are being actively researched by the NLP community. The main challenges with DS-NER are incorrectly labeling terms as nonterms due to the incompleteness of the dictionary and labeling only part of a longer term, such as labeling only "sclerosing cholangitis" in "chronic sclerosing cholangitis" because the dictionary does not contain the longer term. According to our experiments, the former case does not appear to be a serious problem when the training dataset is large; however, the latter case has a considerable impact on the kinds of terms that the model seeks. Ideally, the entire term should be tagged. However, in practice, terms usually appear in multiple layers of nested noun phrases, and with each increasing layer, the essentiality of the term decreases, and where the term should end is subjective. The tagging strategy strongly affects the NER result. For example, when we expanded the tagged term to the longest possible noun phrase containing that term, the trained NER model identified 33 million terms instead of 6 million terms. Thus, the definition of the "right" boundary remains a key problem. For DS-RE, the identification of incomplete entities poses an even greater problem than it does for NER, as it can affect the correctness of the relations. Additionally, for certain relations, the small proportion of sentences (such as 1 out of 10) that actually express the tagged relation can cause the training data to be too noisy to learn the correct sentence patterns.

Ambiguity in evaluation is another difficulty. For the aforementioned nested term boundary problem in NER, instead of being either right or wrong, the boundaries can also be confidently right or less confidently right, which affects all the conventional metrics, such as recall and precision. Ambiguity also affects synonym classification. For example, "cortex" and "cerebral cortex" are technically distinct concepts, but in actual use, "cortex" usually refers to "cerebral cortex", and they are also labeled as synonymous in the UMLS (Concept C0007776). For another example, "coronary artery disease", "coronary heart disease" and "ischemic heart disease" are considered by many to be the same, but they are distinct concepts in the UMLS (Concepts C1956346, C0010068, and C0151744). Because of this ambiguity, a universal classification criterion and objective evaluation are nearly impossible.

While we are confident in our approach for generating BioMedKGs algorithmically, we are not opposed to human intervention. In fact, we are aware that there are many issues in BIOS that may be difficult to solve with models and algorithms alone in the near future, and we will develop a platform to actively interact with users and ask for their contributions. For example, we may detect a pair of terms with a very high similarity according to CODER++ embedding, such as 0.75, and ask users to determine whether the terms are synonymous. In addition, we will actively explore additional sources for knowledge extraction. For example, EHRs could be a good source for identifying nonstandard terms used by doctors in real clinical settings. Certain relations, such as disease-and-drugs and disease-and-laboratory tests, can also be extracted more easily and accurately from EHRs.

## 6. SUMMARY

In this paper, we introduced the methodology for building BIOS, including term discovery, computational identification of concepts, semantic type classification, relation extraction, machine translation, and data and version management. We also conducted a preliminary assessment of the BIOS content. As a BioMedKG built by machine learning from the ground up, BIOS lights the path to future knowledge graph development: not only does it easily reach a gigantic size with very acceptable quality, but it also exhibits significant difference from expert-curated BioMedKGs, such as in term coverage. As machine learning techniques continue to advance and an increasing number of people contribute to the project, we are certain that BIOS will become increasingly accurate and will eventually be larger and more up-to-date than all expert-curated BioMedKGs.